\documentclass[runningheads]{llncs}
\usepackage{multirow}
\usepackage[T1]{fontenc}
\usepackage{amsmath}
\usepackage{enumitem}
\usepackage{graphicx}
\usepackage{subcaption}
\usepackage{booktabs}
\usepackage{tabularx}
\usepackage{xcolor}
\usepackage{multirow}
\usepackage{booktabs}
\usepackage{amsmath} 
\usepackage{xcolor} 
\usepackage{graphicx} 
\usepackage{subcaption} 
\usepackage{amsfonts} 
\usepackage{hyperref}
\usepackage{xcolor}
\begin{document}
\title{S3Simulator: A benchmarking Side Scan Sonar Simulator  dataset for Underwater Image Analysis}
%
%
%
%

\author{kamal Basha~S~$^{1}$ and Athira~Nambiar$^{1}$ }

\date{%
    $^1$SRMIST, Chennai, India \\
    [2ex]%
     \today}

\author{Kamal Basha S\inst{1} and
Athira Nambiar*\inst{1}}
%
%
\institute{Department of Computational Intelligence,\\
Faculty of Engineering and Technology,\\
SRM Institute of Science and Technology\\
Kattankulathur, Tamil Nadu, 603203, India\\ 
\email{c58527@srmist.edu.in, athiram@srmist.edu.in}}

\maketitle   
%
\vspace{-.5cm}
\begin{abstract}
Acoustic sonar imaging systems are widely used for underwater surveillance in both civilian and military sectors. However, acquiring high-quality sonar datasets for training Artificial Intelligence (AI) models confronts challenges such as limited data availability, financial constraints, and data confidentiality. To overcome these challenges, we propose a novel benchmark dataset of Simulated Side-Scan Sonar images, which we term as {\textbf{`S3Simulator dataset'}}. Our dataset creation utilizes advanced simulation techniques to accurately replicate underwater conditions and produce diverse synthetic sonar imaging. In particular, the cutting-edge AI segmentation tool i.e. Segment Anything Model (SAM) is leveraged for optimally isolating and segmenting the object images, such as ships and planes, from real scenes. Further, advanced Computer-Aided Design tools i.e. SelfCAD and simulation software such as Gazebo are employed to create the 3D model and to optimally visualize within realistic environments, respectively. Further, a range of computational imaging techniques are employed to improve the quality of the data, enabling the AI models for the analysis of the sonar images. Extensive analyses are carried out on S3simulator as well as real sonar datasets to validate the performance of AI models for underwater object classification. Our experimental results highlight that the S3Simulator dataset will be a promising benchmark dataset for research on underwater image analysis. \href{https://github.com/bashakamal/S3Simulator}{\textcolor{blue}{https://github.com/bashakamal/S3Simulator}}.

\keywords{Sonar imagery  \and Side Scan Sonar Simulated dataset  \and Segmentation
\and SelfCAD \and Gazebo \and underwater object classification.}
\end{abstract}
\section{Introduction}
SONAR, which stands for Sound Navigation and Ranging, plays a crucial role in various underwater applications~\cite{steiniger2022survey}. Sonar systems utilize sound waves to overcome the limitations posed by optical devices, such as water darkness and turbidity. It has found applications in various civilian and defence sectors. The detection and classification of underwater objects in sonar images remain one of the most challenging tasks in marine applications, such as underwater rescue operations, seabed mapping, and coastal management\textcolor{black}{~\cite{steiniger2022survey}}.

Traditionally, Sonar imagery is manually inspected by human operators, which is a time-consuming task as well as requires domain expertise~\cite{rutledge2018intelligent}. To automate this process, the integration of Artificial Intelligence (AI) emerged as a promising solution. However, the availability of publicly accessible, high-quality sonar datasets to train the AI models is scarce. This paucity of sonar datasets is mainly due to the extensive costs, domain expertise to label, limited resources, security and data sensitivity, and confidentiality constraints. Furthermore, the quality of the available \textcolor{black}{sonar} datasets is also suboptimal due to the complexity of the underwater environment, such as various kinds of distortion, underwater noise, speckle noise, small objects, and poor visibility~\cite{zhang2023detection}.

In order to address the aforementioned challenges, i.e., the scarcity of publicly available sonar data and low-quality sonar images, we propose a synthetic approach for generating  \textbf{S}ide \textbf{S}can \textbf{S}onar \textbf{s}imulator dataset named as \textbf{"S3Simulator"} dataset. \textcolor{black}{Currently, the S3Simulator dataset consists of 600 images of ships and 600 images of planes, which have been meticulously segmented and simulated to replicate real-world sonar conditions.}  A novel framework that combines an advanced AI segmentation model, i.e., Segment Anything Model (SAM)~\cite{kirillov2023segment}, with the selfCAD computer-aided design tool and the 3D dynamic simulator Gazebo is leveraged for the creation of the S3Simulator dataset. Further, it is augmented with cutting-edge computational imaging techniques to provide a heterogeneous dataset replicating real-world sonar imagery such as highlight, shadow, seafloor reverberation, and other characteristics.

The proposed S3Simulator dataset is developed in five stages: First, silhouette images of military and civilian planes and ships are acquired in their raw format. In the second stage, a benchmarking segmentation model SAM is utilized to explicitly segment out the image object, i.e., shipwrecks, plane wreckage, and its fragments, from the rest of the image based on the provided prompts. In the third stage, we employ a self-CAD tool to reconstruct 2D segmented images into 3D models, adjusting the model's properties such as shape, size, and texture. In the fourth stage, these 3D objects are deployed on a simulator platform, e.g., Gazebo, to generate a simulated replica of the real-world objects by rendering the self-CAD results. This environment simulates complex sonar characteristics such as noise, shadows, object complexity, and diverse seabed terrain. Finally, we employ a range of computational imaging techniques, including pixel value clipping, linear gradient integration, and nadir zone mask generation, to enhance the data quality and replicate the characteristics of sonar images.

Extensive analysis is carried out on real and S3Simulator datasets to validate the performance of the AI model for underwater image analysis. In particular, we investigate the application of AI models for sonar image classification. To this end,  benchmarking classical Machine Learning (ML) approaches such as Support Vector Machine (SVM), Random forest, K-Nearest Neighbors (KNN) and  Deep Learning (DL) models including VGG16, VGG19, MobileNetV2, InceptionResNetV2, InceptionV3, ResNet50, and DenseNet121 are trained using augmentation and transfer learning techniques and tested on unseen real data. The key contributions of the paper are as follows:

\begin{itemize}[label=$\bullet$]
    \item Proposal of a novel \textbf{`S3Simulator dataset'} that consists of simulated side-scan sonar imagery to tackle the scarcity of publicly available sonar data and low-quality sonar images.
    
    \item Integration of Gazebo simulator and selfCAD 3D with the advanced AI segmentation model SAM, augmented by computational imaging refining.
    
    \item Incorporation of a realistic environment comprising images with nadir zone, shadows and object rendering, alongside diverse seabed compositions. 


    \item Extensive evaluation of AI models through classical ML and DL methodologies for sonar image classification in both real-world and simulated scenarios.
\end{itemize}

The rest of the paper is organized as follows. The related works are described in Section~\ref{sec:Related Works}. The overall architecture of the proposed S3Simulator multi-stage approach and methodology is presented in Section~\ref{sec:methodology}. In Section~\ref{sec:Sonar image classification on S3Simulator dataset} and Section~\ref{sec:Experimental Setup}, sonar image classification on the S3Simulator dataset and the experimental setup are discussed. In Section~\ref{sec:Experimental Results}, experimental results are discussed in detail. Finally, the conclusion and future works are enumerated in Section~\ref{Conclusion and Future Works}.

\vspace{-.4cm}
\section{Related Works}
\label{sec:Related Works}
\subsection{Sonar Image Dataset}
\label{sec:dataset}
In the exploration of marine and object detection, researchers have made notable progress in creating sonar image datasets. In one of the earliest studies of side-scan sonar datasets, Huo et al.~\cite{huo2020underwater} developed Seabed Objects-KLSG, a side-scan sonar dataset obtained from real sonar equipment, featuring 385 wrecks and 36 drowning victims, 62 airplanes, 129 mines, and 578 seafloor images. Sethuraman et al.~\cite{sethuraman2024machine} AI4Shipwrecks dataset comprises 286 high-resolution side-scan sonar images obtained from autonomous underwater vehicles (AUVs) and labeled with consultation from specialist marine archaeologists. Another dataset i.e. Sonar Common Target Detection Dataset (SCTD)~\cite{zhang2021self} consists of 57 images of planes, 266 images of shipwrecks, and 34 images of drowning victims, each with different dimensions.

However, due to real-world dataset limitations, synthetic sonar images play a significant role in advancing research in underwater exploration. Shin et al.~\cite{shin2022synthetic} proposed a method using the Unreal Engine (UE) to generate synthetic sonar images with various seabed conditions and objects like cubes, cylinders, and spheres. Sung et al.~\cite{sung2020realistic} synthesized realistic sonar images using ray tracing algorithms and GAN. Liu et al.~\cite{liu2021cyclegan} proposed cycle GAN-based generation of realistic acoustic datasets for forward-looking sonars. Yang et al.~\cite{yang2023side} proposed a side-scan sonar image synthesis method based on the diffusion model. Xi et al. ~\cite{xi2024sonar} used optical data to train their developed sonar-style image. Lee et al. ~\cite{lee2018deep} simulated a realistic sonar image of divers by applying the StyleBankNet image synthesizing scheme to the images captured by an underwater simulator.

\subsection{Sonar Image Classification}

After the extensive development of sonar imaging technology, underwater image classification has emerged as a crucial area in the field of ocean development. Li et al.~\cite{li2013research} used Support Vector Machine (SVM) as the classifier to recognize small diver from dim special diver targets accurately and selected five main characteristics of divers such as divers average scale, velocity, shape, direction, and angle with 94.5\% as accuracy rate. After feature extraction, Karine et al.~\cite{karine2015sonar} implemented the k-nearest neighbor (KNN) and SVM algorithms for seafloor image classification recorded by side scan sonar. Zhu et al.~\cite{zhu2017pca} proposed an extreme learning machine (KELM) and principle component analysis (PCA) for side scan sonar image classification. Du et. al~\cite{du2023comparative} compared different CNN model prediction accuracy and found less improvement for AlexNet and VGG-16 and good improvement for Google Net and ResNet101 after the transfer learning technique is applied. Google Net has the highest prediction accuracy at 94.27\%. After fine-tuning limited data, \cite{chungath2023transfer} used a pre-trained deep neural network in which ResNet-34 and DenseNet-121 were the best-performing models of underwater image classification.

In contrast to the aforementioned synthetic sonar images, which are expensive and time-consuming due to recreating from real data, our S3Simulator dataset is quite an economical and time-efficient solution built on simulator technology and advanced AI techniques. To the best of our knowledge, the S3Simulator dataset is the first publicly accessible and extensive compilation of side-scan sonar images for ship and plane objects.
\vspace{-0.4cm} %

\vspace{-.001cm}
\section{S3Simulator Dataset}
\label{sec:methodology}
\begin{figure}[t!]
    \centering 
    \includegraphics[width=1\textwidth]{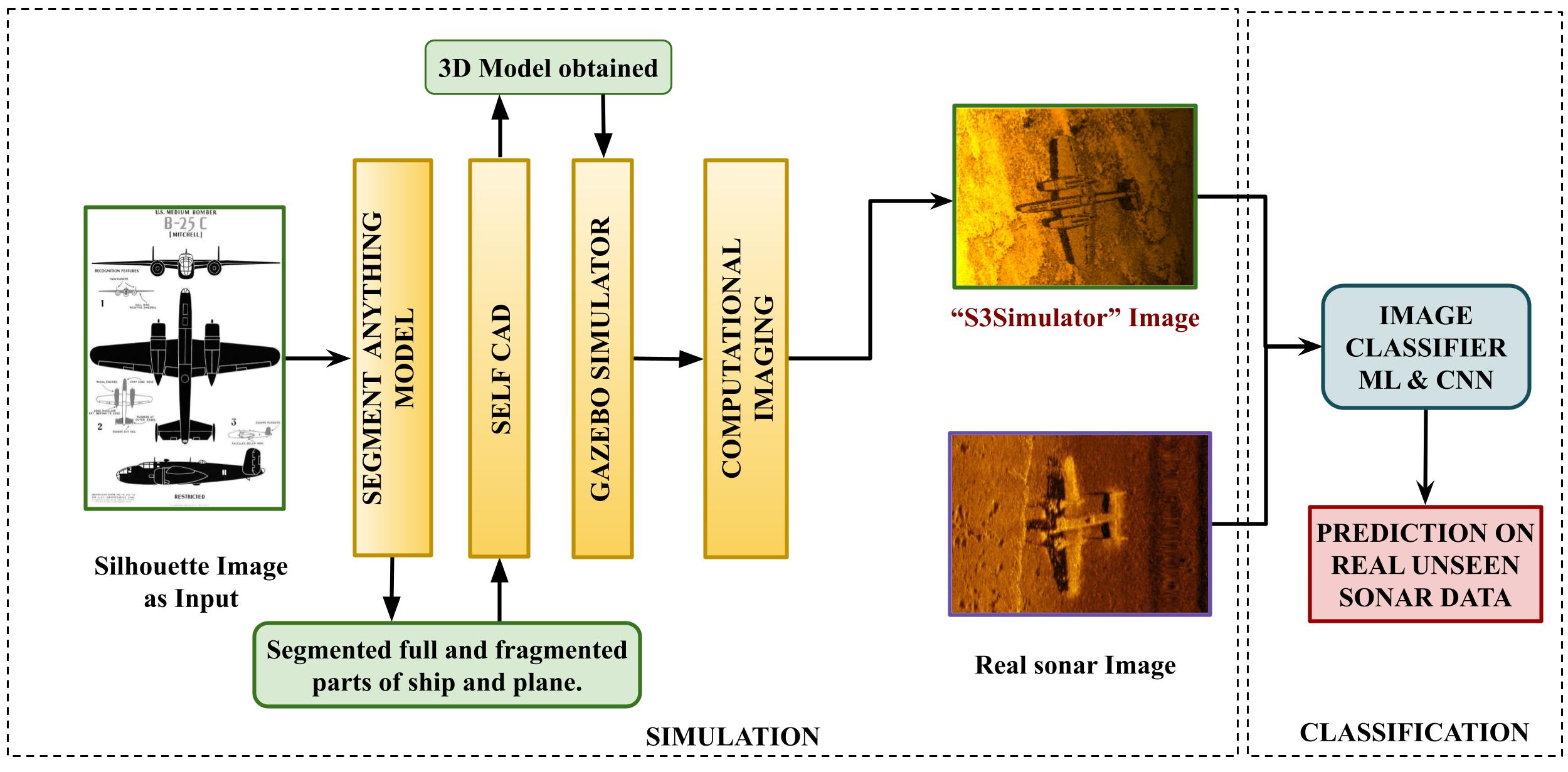}
    \caption{Overall architecture of proposed S3Simulator-based Sonar Image classification}
    \label{fig:architecture updated}
    \vspace{-0.8cm}
\end{figure}

\vspace{-0.2cm} %
This section explains the workflow and generation of the S3Simulator dataset. The overall architecture of the proposed pipeline of S3Simulator is depicted in Fig.~\ref{fig:architecture updated}. It consists of modules Segment Anything Model (SAM), SelfCAD, Gazebo, computational imaging, output of the simulated image, real sonar image, and its classification using Machine Learning (ML) and Deep Learning (DL) techniques. The details of the modules are explained below.
\vspace{-0.26cm} %

\vspace{0.0001cm} %

\subsection{Data acquisition}
\label{dataacquisition}
\vspace{-0.15cm} %
To replicate the sonar imagery of realistic objects such as ships and planes, the 
data is collected from the Royal Observer Corps Club's third-grade exam. The collection has a total of 62 unique aircraft, whereas each aircraft is depicted as a black image on both sides and plan perspectives~\cite{us_navy_ship_silhouettes}, and the U.S. ship silhouettes show the relative size of the various classes of aircraft carriers, battleships, cruisers, and destroyers~\cite{misc1}.{(The images are represented in the supplementary material for reference.)}

Further, for the AI investigation and classification, as mentioned in Section~\ref{sec:dataset} Seabed object KLSG dataset is utilized. \textcolor{black}{(Sample images of the Seabed object KLSG dataset are given in the supplementary material for reference.)} This dataset serves as the basis for testing the S3Simulated dataset against real sonar data.

\subsection{Segmentation with Segment Anything Model (SAM)}
\label{Subsec:SAM}
\begin{figure}[htbp]
    \centering
    \begin{subfigure}{0.23\textwidth}
        \centering
        \includegraphics[width=\textwidth]{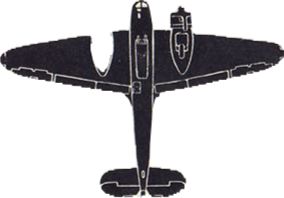}
        \caption{Plane wreck 1}
        \label{fig:sub1}
    \end{subfigure}%
    \begin{subfigure}{0.23\textwidth}
        \centering
        \includegraphics[width=\textwidth]{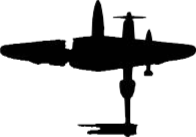}
        \caption{Plane wreck 2}
        \label{fig:sub2}
    \end{subfigure}%
    \begin{subfigure}{0.23\textwidth}
        \centering
        \includegraphics[width=\textwidth]{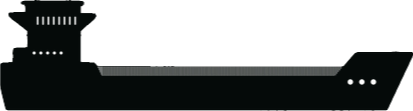}
        \caption{Shipwreck 1}
        \label{fig:sub3}
    \end{subfigure}%
    \begin{subfigure}{0.23\textwidth}
        \centering
        \includegraphics[width=\textwidth]{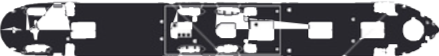}
        \caption{Shipwreck 2}
        \label{fig:sub4}
    \end{subfigure}
    \caption{The Segment Anything Model is utilized to segment fragmented images of ships, aircraft, and vessels from the image.}
    \label{fig:segmentanything}
    \vspace{-.5cm}
\end{figure}

Segment Anything Model (SAM) ~\cite{kirillov2023segment}- is one of the cutting-edge models in semantic segmentation. SAM is intended to identify and isolate an object of interest within an image in response to specific user-provided prompts. Prompts can be text, a bounding box, a collection of points (including a complete mask), or a single point. Even though the request is ambiguous, the model still generates an appropriate segmentation mask, as shown in Fig.~\ref{fig:segmentanything}. Consequently, it can perform effectively in the zero-shot learning regime, i.e. it can segment objects of types it has never encountered before without the need for further training. SAM consists of an image encoder, a flexible prompt encoder, and a fast mask decoder based on Transformer vision models. The image encoder is applied once per image before prompting the model. Masks consist of dense prompts encoded with convolutions and combined element-wise with the image embedding. Image, prompt, and output token embedding are efficiently mapped to masks via the mask decoder.

In the analysis of real-world objects, SAM is applied to facilitate the segmentation and masking of specified objects. In this work, SAM approach is used to segment the planes and ships objects from the raw silhouette images, as shown in Fig.~\ref{fig:segmentanything}.

\subsection{3D Model Generation in SelfCAD}
\label{Subsec:3D Model Generation in SelfCAD}

\begin{figure}[htbp]
    \centering
    \begin{subfigure}{0.23\textwidth}
        \centering
        \includegraphics[width=\textwidth]{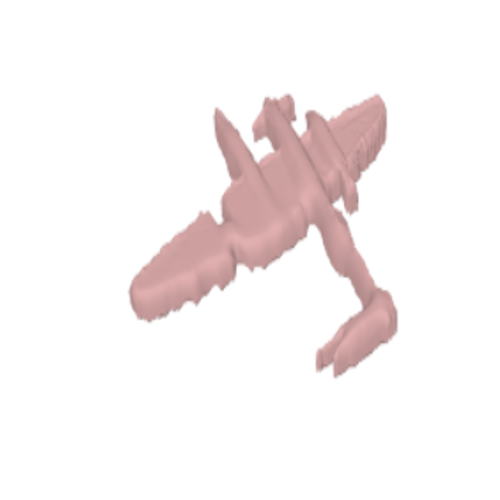}
        \caption{Plane wreck 1}
        \label{fig:sub1}
    \end{subfigure}%
    \begin{subfigure}{0.23\textwidth}
        \centering
        \includegraphics[width=\textwidth]{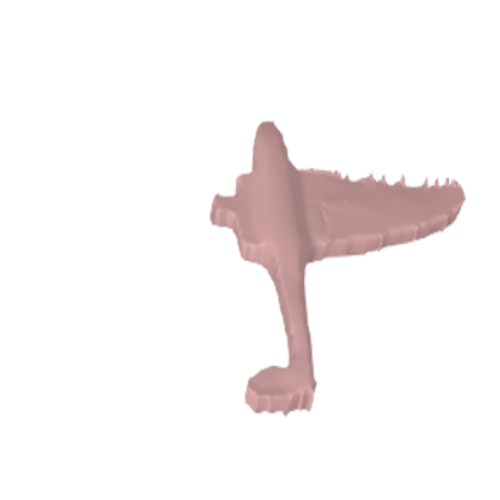}
        \caption{Plane wreck 2}
        \label{fig:sub2}
    \end{subfigure}%
    \begin{subfigure}{0.23\textwidth}
        \centering
        \includegraphics[width=\textwidth]{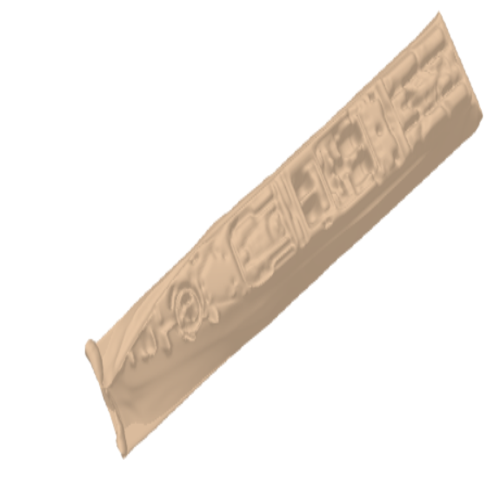}
        \caption{Shipwreck 1}
        \label{fig:sub3}
    \end{subfigure}%
    \begin{subfigure}{0.23\textwidth}
        \centering
        \includegraphics[width=\textwidth]{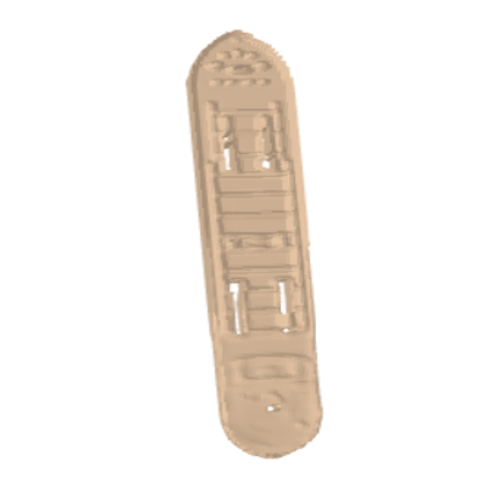}
        \caption{Shipwreck 2}
        \label{fig:sub4}
    \end{subfigure}
    \caption{SelfCAD 3D objects after segmentation}
    \label{fig:selfcad}
        \vspace{-1.2cm}
\end{figure}

SelfCAD ~\cite{selfCAD} is a software application for computer-aided design (CAD) that enables users to modify pre-existing designs as well as to generate 3D model from 2D image. SelfCAD enables users, with its robust tools, to effortlessly create, sculpt, and slice objects. In our work, SelfCAD is employed to generate a 3D model from the segmented 2D images shown in Fig.\ref{fig:segmentanything}. We refined the 3D models by applying sculpting techniques, improving resolution, modifying tolerances, and manipulating size and shape. The purpose of these modifications is to improve and optimize the final 3D models, which are similar to real-world objects as shown in Fig.~\ref{fig:selfcad}.
\vspace{-.8cm}

\subsection{Deployment to the Gazebo Simulator} 
\label{Subsec:Deployment to Gazebo Simulator}

\begin{figure}[htbp]
    \vspace{-.7cm}
    \centering
    \includegraphics[width=0.9\textwidth]{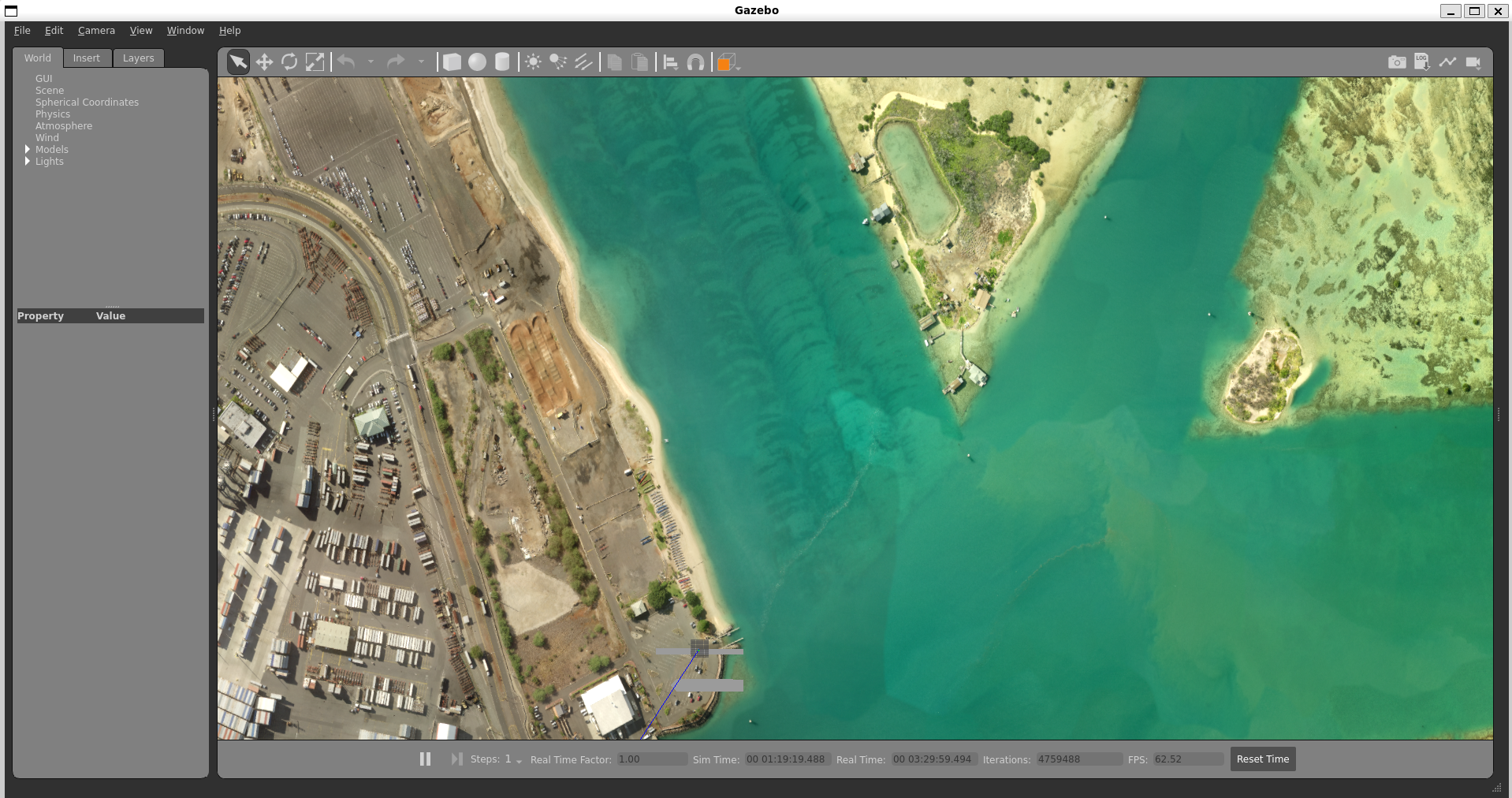}
    \caption{Gazebo environment}
    \label{fig:gazebo environment}
    \vspace{-.5cm}
\end{figure}

\begin{figure}[htbp]
    \centering
    \begin{subfigure}{0.3\textwidth}
        \centering
        \includegraphics[width=0.65\textwidth]{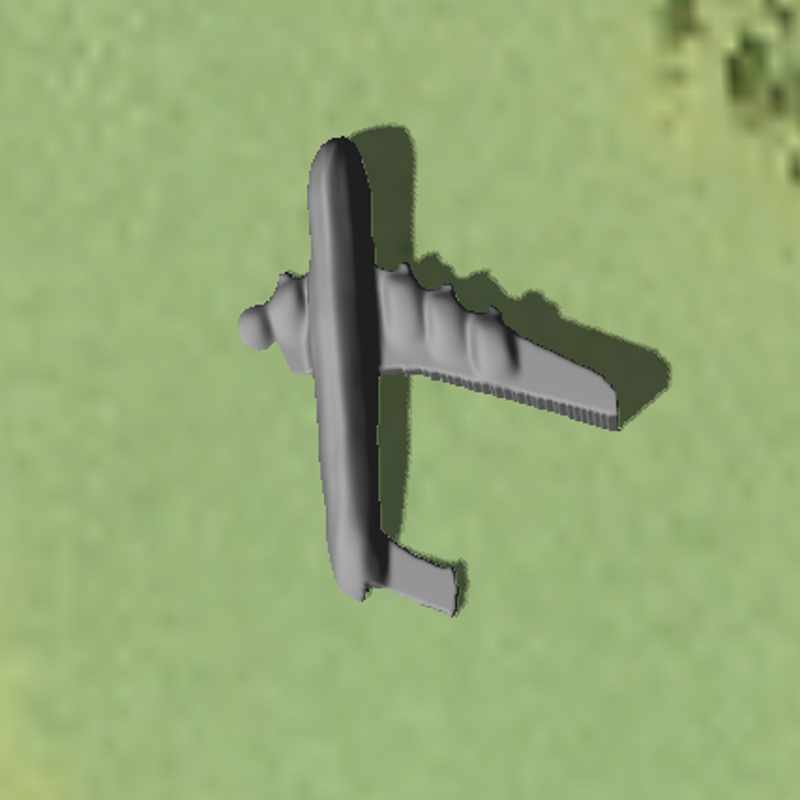}
        \caption{Object rendering}
        \label{fig:sub1}
    \end{subfigure}%
    \hspace{0.3em}
    \begin{subfigure}{0.3\textwidth}
        \centering
        \includegraphics[width=0.65\textwidth]{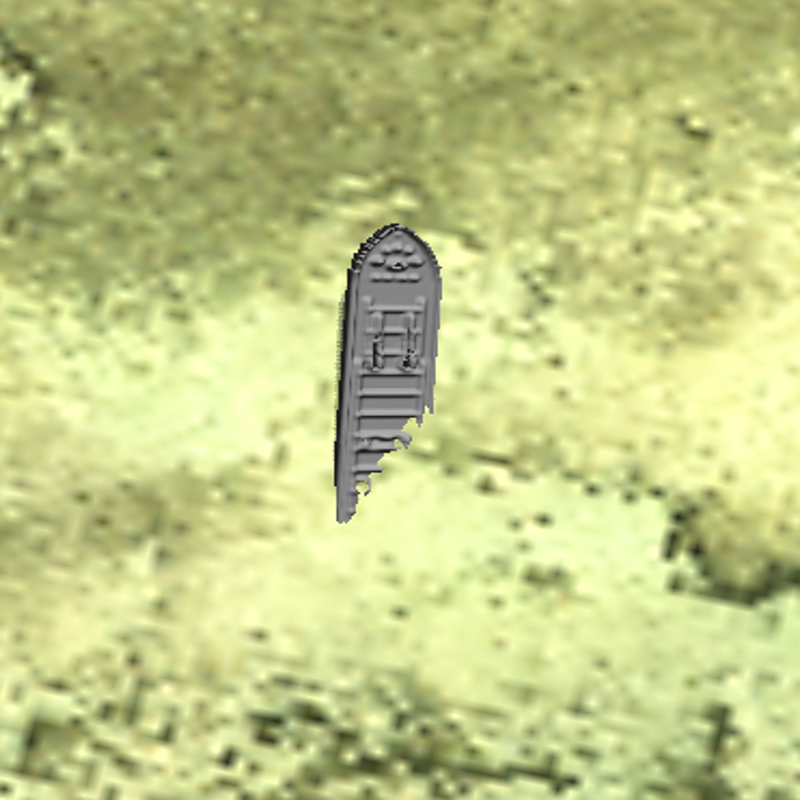}
        \caption{Object rendering}
        \label{fig:sub3}
    \end{subfigure}%
    \hspace{0.3em}
    \begin{subfigure}{0.3\textwidth}
        \centering
        \includegraphics[width=0.65\textwidth]{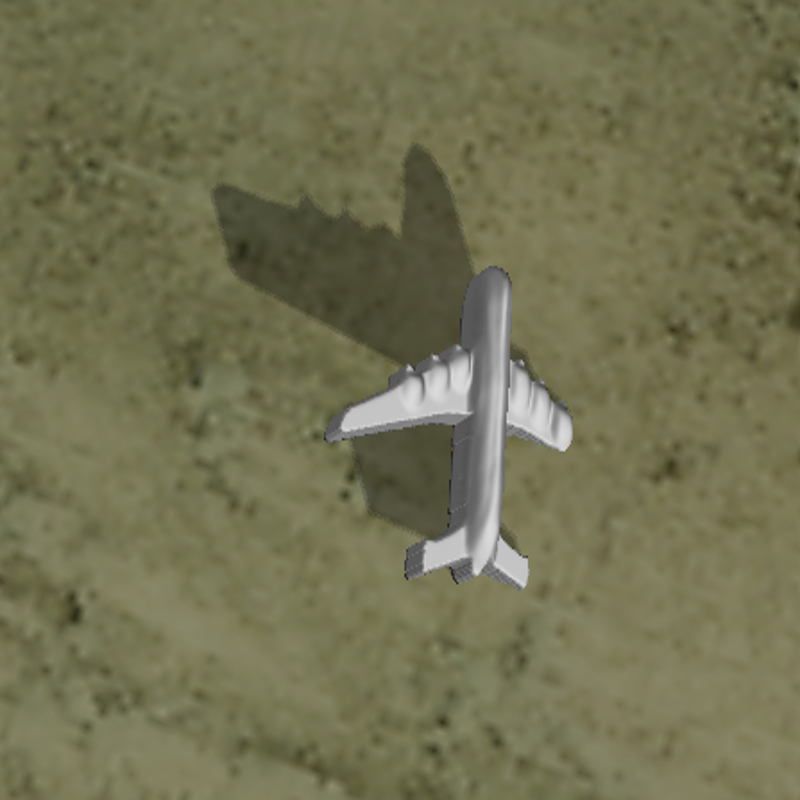}
        \caption{Shadow rendering}
        \label{fig:sub1}
    \end{subfigure}%

    \vspace{0.5em} 

    \begin{subfigure}{0.3\textwidth}
        \centering
        \includegraphics[width=0.65\textwidth]{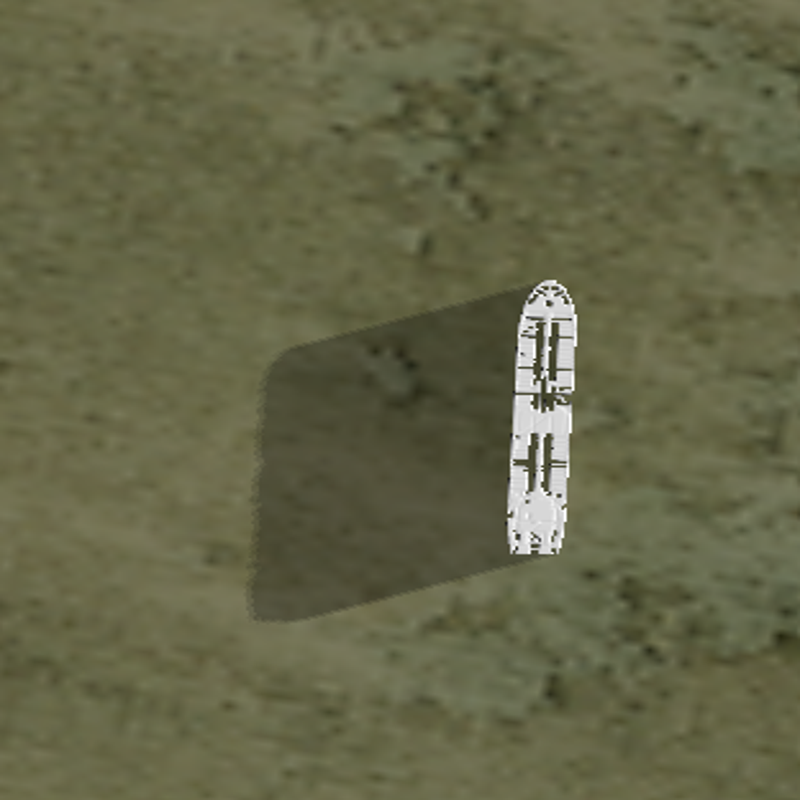}
        \caption{Shadow rendering}
        \label{fig:sub3}
    \end{subfigure}%
    \hspace{0.3em}
    \begin{subfigure}{0.3\textwidth}
        \centering
        \includegraphics[width=0.65\textwidth]{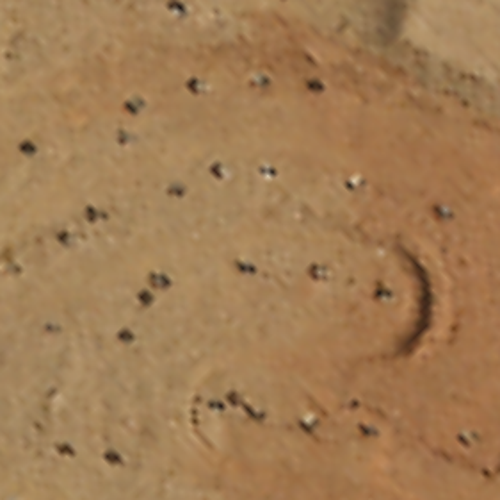}
        \caption{Seabed-1}
        \label{fig:sub1}
    \end{subfigure}%
    \hspace{0.3em}
    \begin{subfigure}{0.3\textwidth}
        \centering
        \includegraphics[width=0.65\textwidth]{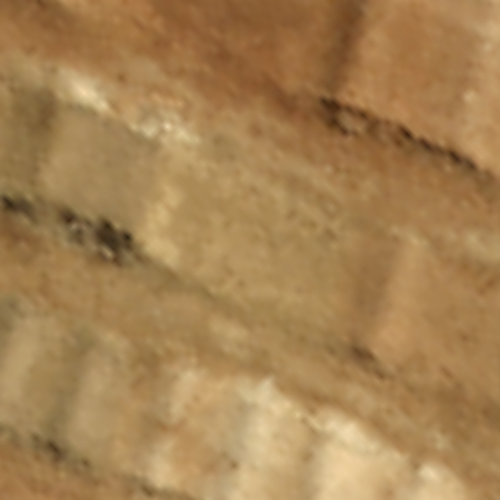}
        \caption{Seabed-2}
        \label{fig:sub2}
    \end{subfigure}%

    \caption{3D object simulated in Gazebo with object rendering (a) and (b), shadow rendering (c) and (d) of plane and ship in different seabed (e) and (f).}
    \label{fig:gazebo_final}
\end{figure}

Gazebo ~\cite{GAZEBO} is an open-source robotic simulator that simplifies high-performance application development. Its primary users are robot designers, developers, and educators. In our work, Gazebo is employed to simulate sonar images by rendering 3D objects and shadows on various seabeds shown in Fig.~\ref{fig:gazebo_final}. The generated 3D model is integrated into Gazebo World. The rendering of objects is achieved by adjusting their poses on the x, y, and z axes and incorporating features like roll, pitch, and yaw rotations. Additionally, the visual texture of the 3D model can be fine-tuned with RGB values from the link inspector available in the Gazebo. To bring the simulated image to a more realistic sonar image, we explicitly add noise from sensors provided by Gazebo models, which adds Gaussian-sampled disturbance independently to each pixel.

\vspace{-0.1cm} 
\subsection{Computational Imaging}
\label{Subsec:Image processing}

This process includes a series of computational imaging techniques aimed at converting data from a Gazebo simulator into a visual representation that closely resembles sonar imagery. Some of the key techniques include the clipping of pixel values, the integration of linear gradients, and the generation of nadir zones.
\vspace{-.4cm}
\subsubsection{Image clipping and integration of Linear Gradient}
\label{Subsec:Integration of Linear Gradient}
\begin{figure}[t!]
    \centering
    \begin{subfigure}{0.24\textwidth}
        \centering
        \includegraphics[width=\textwidth]{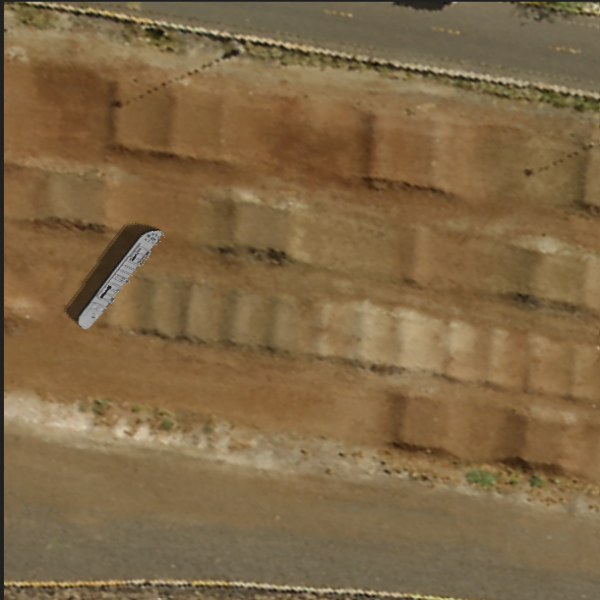}
        \caption{Simulated Image}
        \label{fig:sub1}
    \end{subfigure}%
    \hspace{0.5cm}
    \centering
    \begin{subfigure}{0.34\textwidth}
        \centering
        \includegraphics[width=\textwidth]{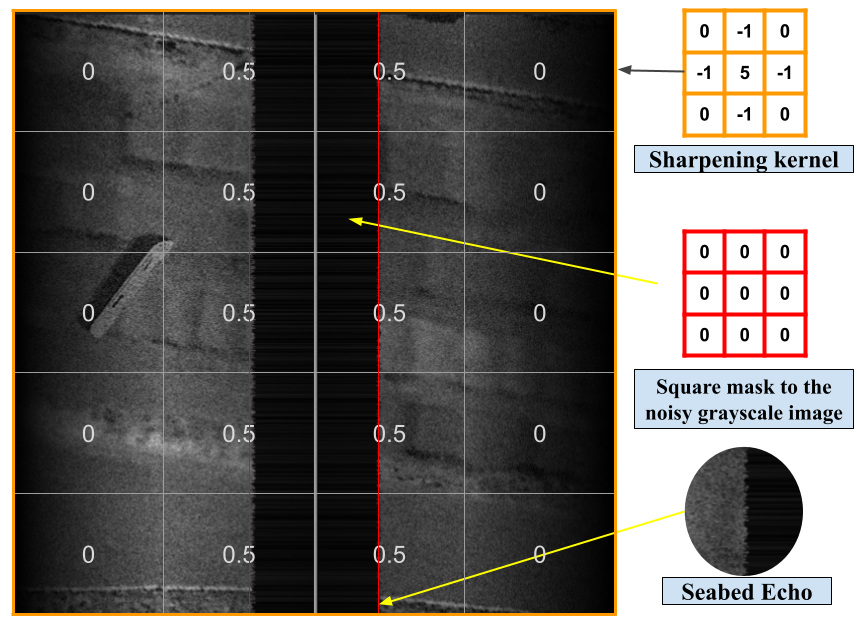}
        \caption{Nadir Zone}
        \label{fig:sub1}
    \end{subfigure}%
    \hspace{0.5cm}
    \begin{subfigure}{0.24\textwidth}
        \centering
        \includegraphics[width=\textwidth]{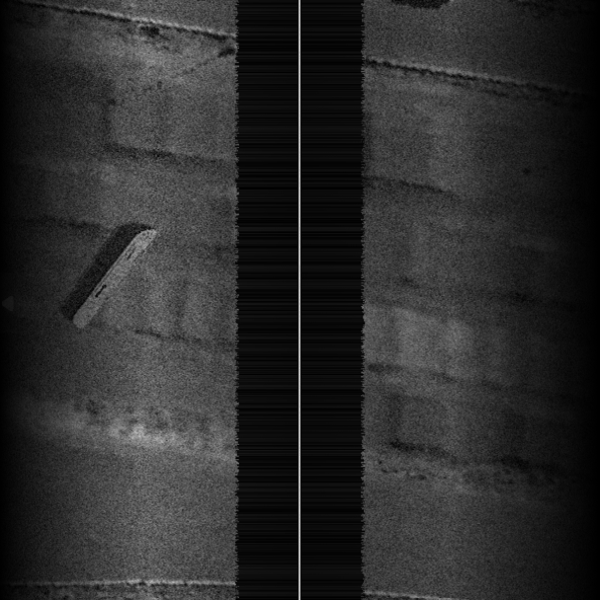}
        \caption{Output}
        \label{fig:sub2}
    \end{subfigure}%
     \caption{Image (a) represents the simulated image from the gazebo; (b) represents the integration of the linear gradient and nadir zone; and (c) represents the final output image after the computational imaging technique.}
      \label{fig:Computational imaging}
\end{figure}

In a sonar image, we can identify that dark colours represent deeper areas and bright colours represent shallow areas. To mimic the real-world conditions, the linear gradient technique is employed in simulated sonar images by partitioning the image into 50\% and applying a gradient on both sides as shown in fig ~\ref{fig:Computational imaging}. 
The gradient for the image function is given by: 
\vspace{-.3cm}
\begin{equation}
\Delta I = \left[ \frac{\partial I}{\partial x}, \frac{\partial I}{\partial y} \right]
\end{equation}

Gradient for Quarter-based Intensity Mapping in a Simulated Sonar Image,
\begin{equation}
\Delta I(x, y) = \begin{cases}
0 & \text{if } x \leq 0.25 \cdot w \text{ (first quarter)} \\
0.5 & \text{if } 0.25 \cdot w < x \leq 0.5 \cdot w \text{ (second quarter)} \\
0.9 & \text{if } 0.5 \cdot w < x \leq 0.75 \cdot w \text{ (third quarter)} \\
0.5 & \text{if } 0.75 \cdot w < x \leq w \text{ (last quarter)}
\end{cases}
\end{equation}

In this representation:
\begin{itemize}
  \item \( w \) - width of the image.
  \item \( \Delta I(x, y) \) - gradient intensity at position \( (x, y) \) in the image.
  \item The gradient changes at different rates depending on the value of \( x \), where \( x \) is the horizontal position within the image.
  \item The gradient is 0 in the first quarter of the width, 0.5 in the second, 0.9 in the third, and 0 in the last.

\end{itemize}

\vspace{-1cm}
\subsubsection{Generation of nadir zone} 
\label{Subsec:Nadir}
The nadir zone in the sonar image is beneath the sonar sensor which appears as a dark zone with a thick white line in between the zone. To mimic this in our work the combination of clipping and linear gradient is employed, to mask the nadir zone and thin white line inside the zone. 

\begin{equation}
K(x, y) = \begin{cases} 
1 & \text{if } I(x, y) \geq Th \\
0 & \text{otherwise}
\end{cases}
\end{equation}

where:
\begin{itemize}
    \item \( Th \) - chosen threshold value.
    \item \( K(x, y) \) - resulting mask, with a value of 1 indicating that the pixel \( (x, y) \) is part of the Nidar zone and 0 indicating it's not.
\end{itemize}


\subsection{Image Augmentation}
\label{Subsec:Image Augmentation}
\vspace{-0.26cm} %

Image augmentation techniques~\cite{xu2023comprehensive} are utilized to improve the resilience and classification performance of the models, trained on both the synthetic sonar simulation dataset and the real sonar dataset. The augmentation pipeline consists of modifications to horizontal flips and random crops. These changes facilitate the inclusion of alterations in the images, hence strengthening the model's ability to acquire robust features and mitigating the issue of overfitting.


\section{Sonar image classification on S3Simulator dataset}
\label{sec:Sonar image classification on S3Simulator dataset}
\vspace{-0.26cm} %
To showcase the effectiveness of the proposed S3Simulator, the developed dataset is used to benchmark computer vision applications such as image classification. Image classification is a fundamental task in computer vision that involves categorizing an image into one or more predefined classes~\cite{lai2019comparison}. This study explores two primary techniques for image classification: the classical Machine Learning (ML) approach, which utilizes algorithms like k-nearest neighbors (KNN), Random Forest, and Support Vector Machines (SVM), suitable for smaller datasets; and deep learning (DL) models, such as Convolutional Neural Networks (CNNs), which can automatically learn from huge datasets with intricate patterns to provide better results.

The k-nearest neighbors~\cite{guo2003knn} algorithm is a simple approach that classifies new data by calculating the distance between the nearest neighbors. similarly, Random forest~\cite{inbook} is a technique in ensemble learning that combines predictions from multiple decision trees that were trained on random subsets of data. Whereas, SVM~\cite{li2013research} utilize a hyperplane in the feature space to distinguish between classes. Formally, the aforementioned techniques can be represented as 

\vspace{-0.25cm}
\begin{equation}
    f(x) = \sum_{i}^n \alpha_i y_i K(x, x_i) + b
    \label{svm}
\end{equation}
\textcolor{black}{where, $f(x)$ is the predicted label for the input $x$, $\alpha_i$ is lagrange multipliers obtained from the SVM optimization process, $x_i$ the feature vector of the training data, $y_i$ is class labels (-1 or 1) for the training data point, $K(x,x_i)$ represents the kernel function that calculates the similarity between the input vector $x$ and the support vectors $x_i$, and $b$ is the bias to determine the offset of the decision boundary.}

Deep Learning (DL) is widely used in the field of pattern recognition and are more efficient than traditional machine learning approaches for image classification~\cite{o2015introduction}. In particular,  Convolutional Neural Networks (CNN) are utilized to classify images. In our study, we leverage transfer learning approach~\cite{weiss2016survey} wherein knowledge from one model is transferred to another model, in order to train deep neural networks with comparatively little data. Mathematically, the neural network learning can be represented as follows.

\vspace{-0.25cm}
\begin{equation}
f = \sigma \left( W_n * h + b_n \right)
\end{equation}

\noindent where, $f$ is the predicted label, $\sigma$ be the activation function, $W_n$ is the weights of the newly added fully-connected layer for binary classification, h is the Output from the pre-trained model (usually the last layer before the final classification layer in the backbone model) and $b_n$ is the biases of the newly added fully-connected layer.

\vspace{-.6cm}
\section{Experimental Setup}
\label{sec:Experimental Setup}
\vspace{-.4cm}
In this Section, the experimental details employed to develop, train, and evaluate the sonar image classification is explained. 
\vspace{-.5cm}

\subsection{Dataset}
\label{sub:datasest}
In the generation of the S3Simulator dataset, a 3D object model is generated from the silhouette images. To enhance realism with realistic objects, the silhouette image is acquired from army fighter planes, army bombers, naval planes, and battleships, as mentioned in Section ~\ref{dataacquisition}.

For AI investigation, we incorporate a simulated dataset S3Simulator along with a real sonar Seabed objects-KLSG dataset~\cite{huo2020underwater}. This dataset comprises 578 seafloor images, 385 wreck images, 36 drowning victim images, 62 aircraft images, and 129 mine images accumulated over a period of more than ten years. With the generous assistance of numerous sonar equipment suppliers—including Lcocean, Hydro-tech Marine, Klein Marine, Tritech, and EdgeTech—this dataset is made possible. Additionally, it comprises images that were obtained directly from the original large-scale side-scan sonar images without any preprocessing.
\vspace{-.5cm}

\subsection{Evaluation protocol}
\label{head:evaluation protocol}

In the evaluation of the sonar image classification task, benchmarking classification metrics such as accuracy and confusion matrix are used. 
Accuracy is an evaluation metric that allows to measure the total number of predictions a model gets right. Mathematically, Accuracy (ACC) is formulated as, 


\begin{equation}
\text{ACC} = \frac{(TP + TN)}{(TP + FP + FN + TN)},
\end{equation}

\begin{itemize}
    \item $\text{TP}$ (True Positives) - number of images correctly classified as positive.
    \item $\text{TN}$ (True Negatives) - number of images correctly classified as negative.
    \item $\text{FP}$ (False Positives) - number of images incorrectly classified as positive.
    \item $\text{FN}$ (False Negatives) - number of images incorrectly classified as negative.
\end{itemize}

Confusion matrix displays counts of the True Positives, False Positives, True Negatives, and False Negatives produced by a model as shown in Fig.~\ref{fig:confsion final}. Using a confusion matrix we can get the values needed to compute the accuracy of a model.
\vspace{-0.5cm}

\subsection{Implementation details}
\label{sub:implementation}

In this investigation, pre-trained models such as VGG16, VGG19, MobileNetV2, InceptionResNetV2, InceptionV3, ResNet50, and DenseNet121 are trained on the ImageNet dataset with two active layers of 1024 and 512 neurons. A dropout layer with a 0.25 dropout rate and a batch normalization layer improved the model's robustness. In training, we used the Adam optimizer with a learning rate of 0.0001 and a batch size of 16. We employed model checkpoint and early stopping during the evaluation to evaluate training progress and prevent overfitting.
In the gazebo, models are described as Simulation Description Format (.sdf) files detailing their physics characteristics, properties, visual appearance, collision, etc. We Utilized Ubuntu 22.04.4 LTS and Gazebo multi-robot simulator, version 11.10.2 for gazebo simulation. The implementation of image classification is conducted on Google Colab, utilizing an T4 GPU with an allocation of 15GB of RAM for training and employing Google’s TensorFlow framework.  
\vspace{-0.5cm}

\section{Experimental Results}
\label{sec:Experimental Results}

\begin{figure}[t!]
       \includegraphics[width=\textwidth]{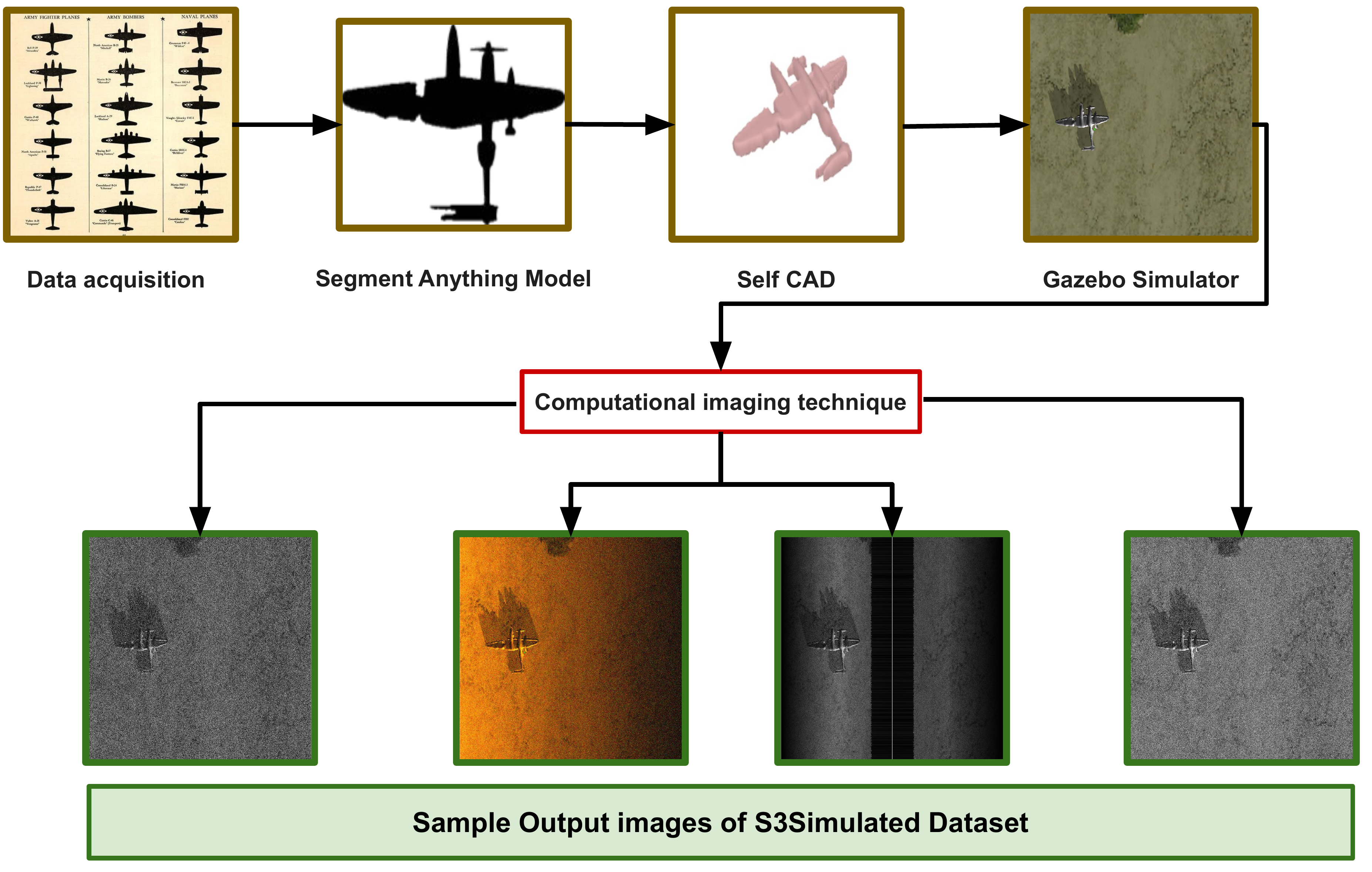}
        \label{fig:sub1}
    \caption{Pipeline of S3Simulated dataset}
    \label{fig:pipeline}
\end{figure}

\subsection{S3Simulator dataset results}
\begin{figure}[h!]
    \centering
    \begin{subfigure}{0.2\textwidth}
        \centering
        \includegraphics[width=\textwidth]{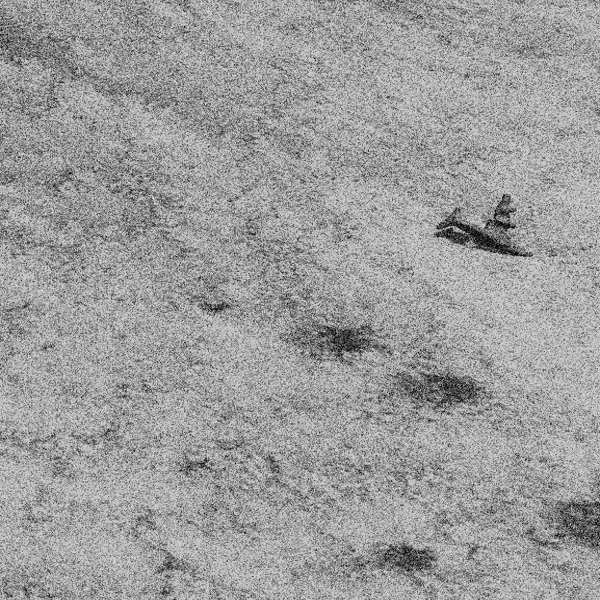}
        \label{fig:sub1}
    \end{subfigure}%
    \hspace{0.5em} 
    \begin{subfigure}{0.2\textwidth}
        \centering
        \includegraphics[width=\textwidth]{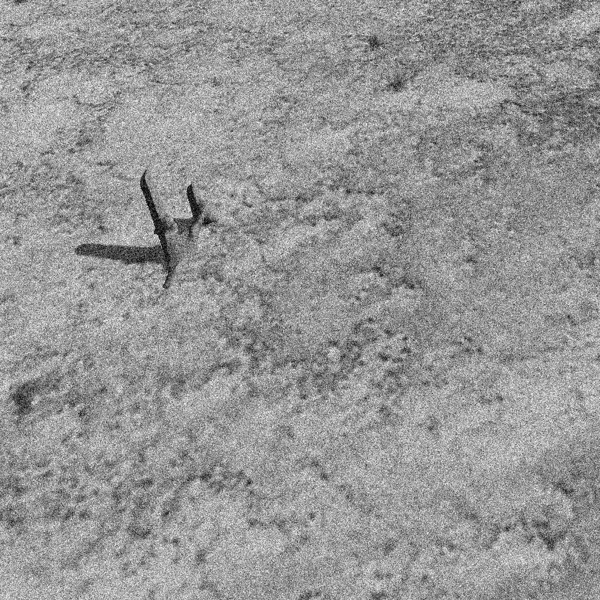}
        \label{fig:sub4}
    \end{subfigure}
    \hspace{0.5em} 
    \begin{subfigure}{0.2\textwidth}
        \centering
        \includegraphics[width=\textwidth]{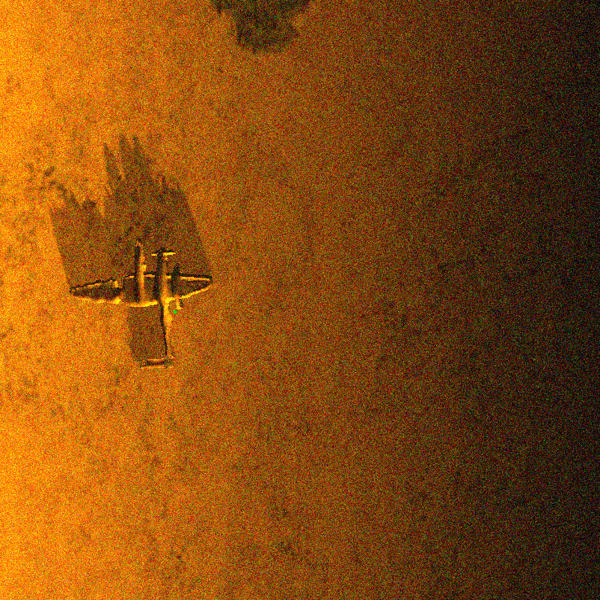}
        \label{fig:sub8}
    \end{subfigure}%
    \hspace{0.5em} 
    \begin{subfigure}{0.2\textwidth}
        \centering
        \includegraphics[width=\textwidth]{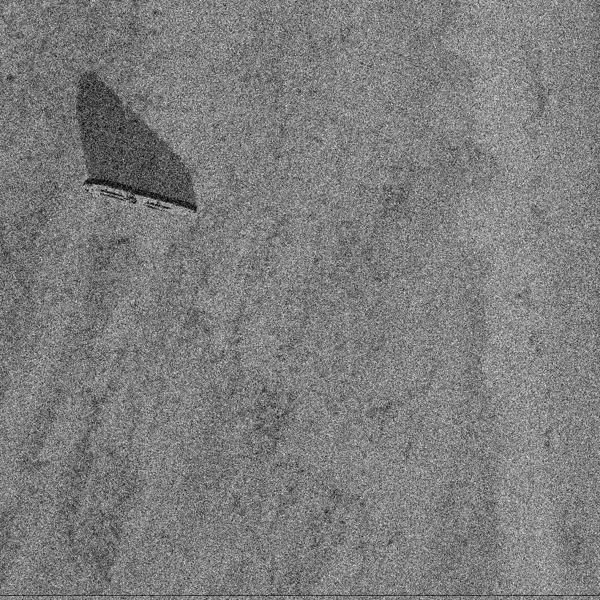}
        \label{fig:sub4}
    \end{subfigure}
    \hspace{0.5em} 
    \begin{subfigure}{0.2\textwidth}
        \centering
        \includegraphics[width=\textwidth]{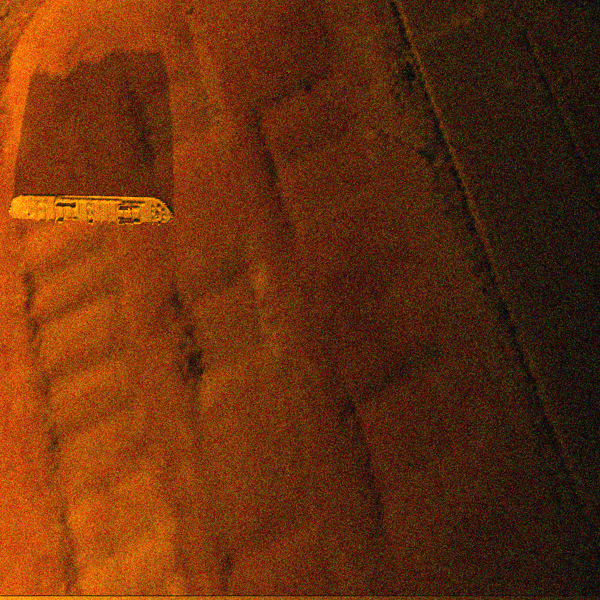}
        \label{fig:sub8}
    \end{subfigure}%
     \hspace{0.5em} 
    \begin{subfigure}{0.2\textwidth}
        \centering
        \includegraphics[width=\textwidth]{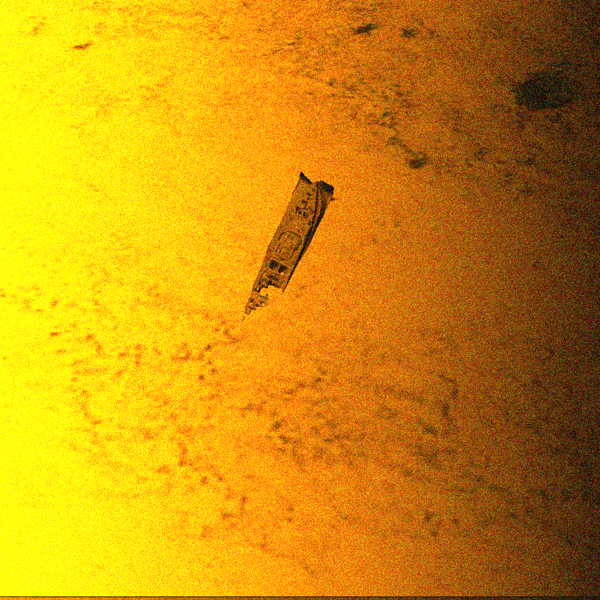}
        \label{fig:sub8}
    \end{subfigure}%
     \hspace{0.5em} 
    \begin{subfigure}{0.2\textwidth}
        \centering
        \includegraphics[width=\textwidth]{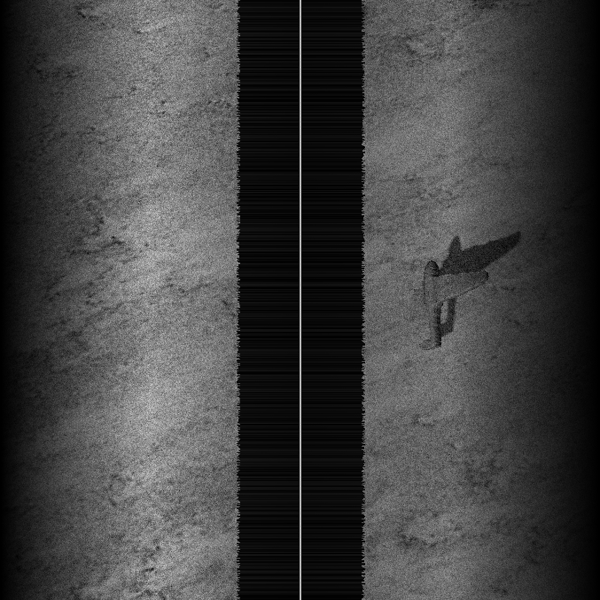}
        \label{fig:sub12}
    \end{subfigure}%
    \hspace{0.5em}
    \begin{subfigure}{0.2\textwidth}
        \centering
        \includegraphics[width=\textwidth]{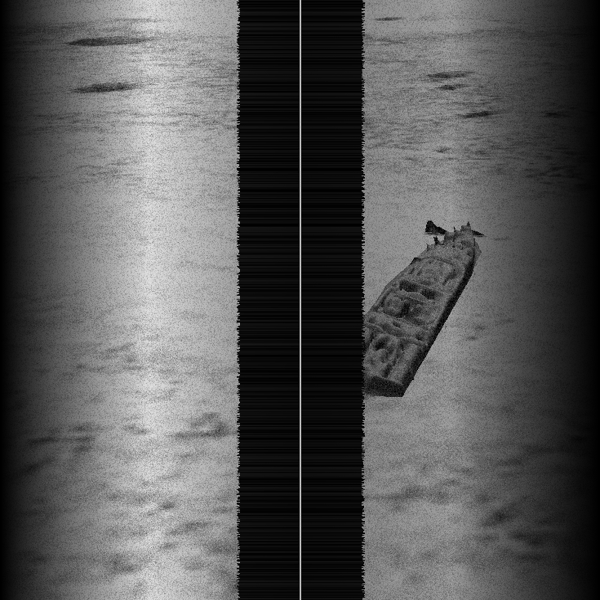}
        \label{fig:sub9}
    \end{subfigure}%
    \caption{S3Simulated Dataset images of ship and plane}
    \label{fig:sample image}
\end{figure}

As explained in Section~\ref{sec:methodology} overall architecture, the final pipeline of simulated sonar image is shown in Fig.~\ref{fig:pipeline}. The pipeline consists of data acquisition, SAM, self-CAD, gazebo simulator, and computational imaging techniques employed to generate the S3Simulator dataset. Sample S3Simulator images shown in Fig.~\ref{fig:sample image}.

\vspace{-.5cm}
\subsection{Sonar Image classification}
Referring to Section~\ref{sec:Sonar image classification on S3Simulator dataset}, a benchmark study of classification using the S3Simulator dataset is carried out using both Machine Learning (ML) and Deep Learning (DL) classifiers. Extensive analysis using simulated data, real data, and real+simulated data is conducted. The classifier performance of ML and DL models are shown in Table~\ref{tab:Classifier_performance} and Table~\ref{tab:DLmodel_performance1}. From Table~\ref{tab:Classifier_performance}, it can be observed that by using simulated data SVM outperforms with an accuracy of 92\%. Similarly, the Random Forest classifier outperforms with a test accuracy of 88\% in real data. While utilizing both real + simulated data and testing on real data, which represents a realistic deployment of scenarios in the wild, SVM classifier outperforms with 88\% accuracy. From all the above analyses, SVM classifier was found to be providing superior performance among all.

\vspace{0cm}
\begin{table}[t!]
    \centering
    \caption{ML Classifier performance of both real and simulated sonar datasets}
    \begin{tabular}{|c|c|c|c|c|}
        \hline
        \textbf{Training Data} & \textbf{Testing Data} & \textbf{Classifier} & \textbf{Train Accuracy} & \textbf{Test Accuracy} \\
        \hline
        \multirow{3}{*}{Simulated} & \multirow{3}{*}{Simulated} & \textbf{SVM} & 1.00 & \textbf{0.92} \\
        & & Random Forest & 1.00 & 0.69 \\
        & & KNN & 0.75 & 0.56 \\
        \hline
        \multirow{3}{*}{Real} & \multirow{3}{*}{Real} & SVM & 1.00 & 0.77 \\
        & & \textbf{Random Forest} & 1.00 & \textbf{0.83} \\
        & & KNN & 0.83 & 0.72 \\
        \hline
        \multirow{3}{*}{Real + Simulated} & \multirow{3}{*}{Real} & \textbf{SVM} & 1.00 & \textbf{0.88} \\
        & & Random Forest & 1.00 & 0.63 \\
        & & KNN & 0.79 & 0.65 \\
        \hline
        \hline
    \end{tabular}
    \label{tab:Classifier_performance}
\end{table}


\begin{table}[th!]
  \centering
   \caption{\textcolor{black}{Test accuracy of different models tested in real data}}
  \begin{tabularx}{\textwidth}{|X|>{\centering\arraybackslash}X|>{\centering\arraybackslash}X|>{\centering\arraybackslash}X|}
    \toprule
    \textbf{Model} & \textbf{Trained in real data} & \textbf{Trained in real + simulated data} & \textbf{Percentage improved from real data to combined data} \\
    \midrule
    VGG\_16 & 0.90 & 0.94 & 4\%\\
    VGG\_19 & 0.87 & 0.92 & 5\%\\
    ResNet50 & 0.64 & 0.70 & 6\%\\
    InceptionV3 & 0.91 & 0.94 & 3\%\\
    DenseNet121 & 0.92 & \textbf{0.96} & 4\%\\
    MobileNetV2 & 0.89 & 0.94 & 5\%\\
    InceptionResNetV2 & 0.91 & 0.95 & 4\%\\
    \bottomrule
    \bottomrule
  \end{tabularx}%
  
   \vspace{-.8cm}
  \label{tab:DLmodel_performance1}
\end{table}%

\textcolor{black}{Analogous to the ML classifier, the performance of the DL classifier, as mentioned in Section~\ref{sub:implementation}, is also investigated. Referring to Table~\ref{tab:DLmodel_performance1}, the test accuracy of different models in test data that are trained using real data / real + simulated data are studied. It is observed that while training with real data, DenseNet121 and InceptionResNetV2 outperform the models with an accuracy of 92\% and 91\%, respectively. Further training with real+simulated data,  DenseNet121 achieved the best performance with test accuracy of 96\%. It is observed that the significant improvement in accuracy of all the models from 3\%-6\% is observed in real + simulated data compared to real data. This accentuates the impact of additional synthetic data augmenting the training process, by replicating realistic sonar data in terms of the number of images, and quality of images and by recreating real-world scenarios.}

\begin{figure}[t!]
    \centering
    \begin{subfigure}{0.55\textwidth}
        \includegraphics[width=\textwidth]{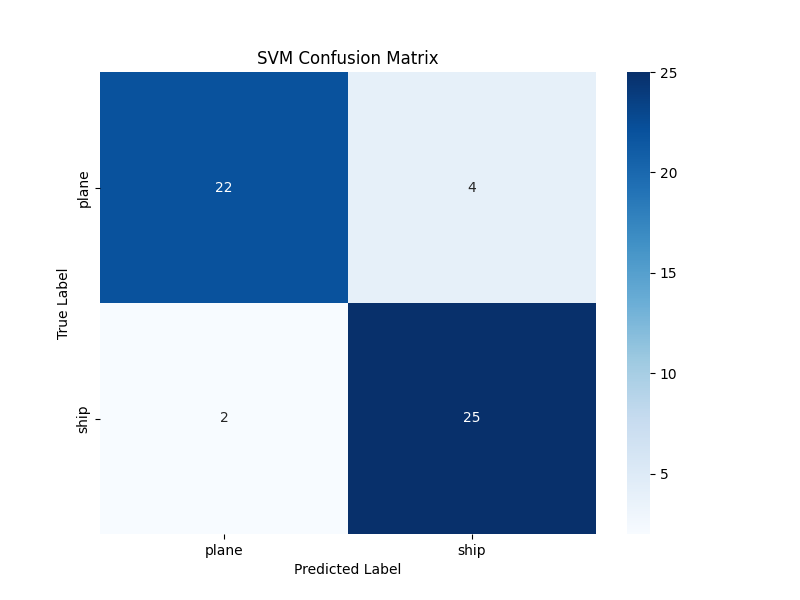}
        \caption{ML classifier-SVM}
        \label{fig:sub1}
    \end{subfigure}%
    \begin{subfigure}{0.46\textwidth}
        \centering
        \includegraphics[width=\textwidth]{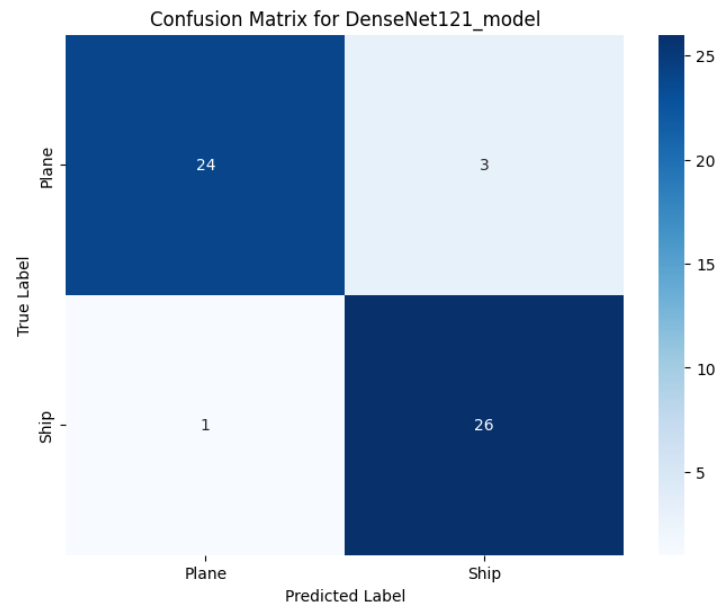}
        \caption{DL Classifier-DenseNet121}
        \label{fig:sub2}
    \end{subfigure}%
\caption{Confusion matrix of the highest-performing classifiers i.e. Support Vector Machine(SVM) and DenseNet121, respectively.}
    \label{fig:confsion final}
\end{figure}

The confusion matrices of the best-performed models in both ML (i.e., SVM) and DL (i.e.,DenseNet121)  are depicted in Fig.~\ref{fig:confsion final}. The overall as well as class-wise accuracy is analysed in the test scenario. The accuracy of the best ML classifier and DL classifier are 88\% and 96\%, respectively. It is observed that the performance of the "plane" class is improved in the DL Model with an accuracy of 88\%  compared to the ML accuracy of 84\%. Similarly, the accuracy of "ships" is increased from 92\% to 96\% in the DL model. 



    

\subsection{Visualization: Highlights \& Shadows}

\begin{figure}[t!]
    \centering
    \begin{subfigure}[t]{0.48\textwidth}
        \centering
        \includegraphics[width=\textwidth]{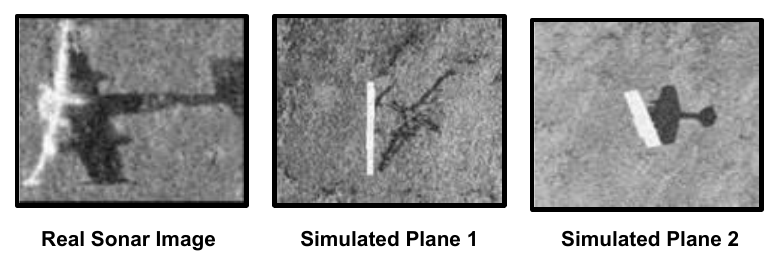}
        \caption{\textcolor{black}{Highlighting the Shadow Characteristics in Real and Simulated Sonar Images}}
        \label{fig:shadow of real and sim}
    \end{subfigure}
    \hfill
    \begin{subfigure}[t]{0.48\textwidth}
        \centering
        \includegraphics[width=\textwidth]{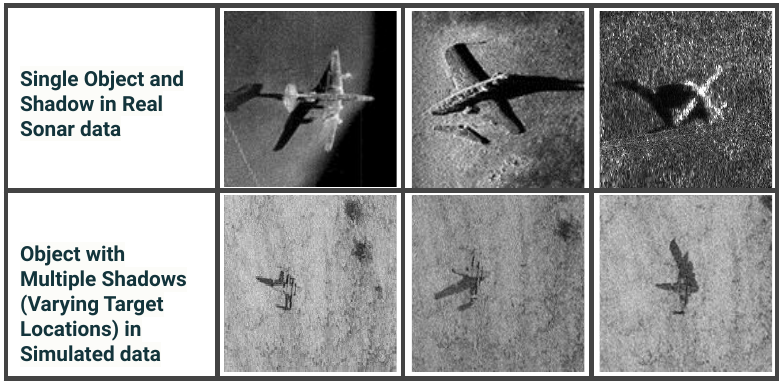}
        \caption{\textcolor{black}{Contrasting Singular and Diverse Shadow Representations in Real vs. Simulated Sonar Imagery}}
        \label{fig:shadow_contrast}
    \end{subfigure}
    \caption{\textcolor{black}{Leveraging Simulated Sonar Shadows to Enhance Real-World Object Identification}}
    \label{fig:combined_images}
\end{figure}

\textcolor{black}{In sonar image analysis, both the shadow and highlight regions are crucial for accurate classification and detection, as they provide complementary information about the objects' shape and size.  In many cases, the highlight area of an object in a sonar image may not be clearly visible, but its shadow can be distinctly observed as shown in Fig.~\ref{fig:shadow of real and sim}. By emphasizing the shadow information, the "S3Simulator" dataset addresses an important gap in existing sonar image datasets and provides a valuable resource for researchers to advance the field of sonar image analysis.}

\begin{figure}[t!]
       \includegraphics[width=\textwidth]{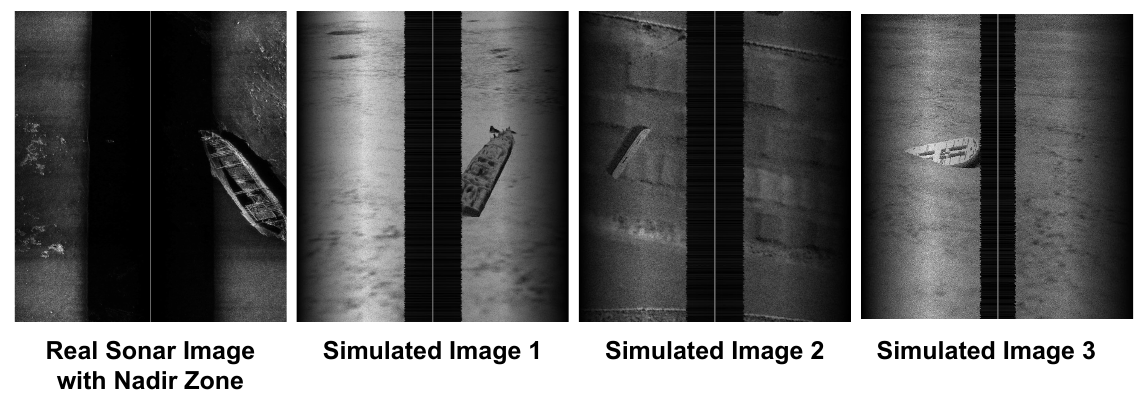}
        \label{fig:sub1}
    \caption{\textcolor{black}{Addressing the Nadir Zone Challenge in Sonar Simulation through the "S3Simulator" Dataset.}}
    \label{fig:nadir_zone}
    \vspace{-.7cm}
\end{figure}

\textcolor{black}{The shadow characteristics in publicly available datasets are typically determined by the fixed positioning of the sonar relative to the object. The "S3Simulator" dataset overcomes this limitation by allowing for the generation of sonar images with varying shadow angles for a given object. This is achieved through the flexibility of the simulation process, where the position and orientation of the sonar device can be adjusted to create images with different shadow characteristics, as shown in Fig.~\ref{fig:shadow_contrast}}. Furthermore, real-world side-scan sonar data contains a nadir zone, a crucial feature often missing in synthetic datasets. The "S3Simulator" dataset uniquely incorporates the nadir zone, enhancing the realism of simulated sonar imagery as shown in Fig.\ref{fig:nadir_zone}.

\section{Conclusion and Future Works}
\label{Conclusion and Future Works}
\vspace{-.2cm}
In this work, we presented a novel benchmarking \textbf{S}ide \textbf{S}can \textbf{S}onar simulator dataset named \textbf{"S3Simulator dataset"} for underwater sonar image analysis. By employing a systematic methodology that encompasses the collection of real-world images, reconstruction of 3D models, simulations, and computational imaging techniques, we have effectively generated a comprehensive dataset similar to real-world sonar images. The effectiveness of our methodology is demonstrated by benchmarking image classification results obtained from several machine learning and deep learning techniques applied to both simulated and real sonar datasets. Future enhancements to our approach aim to increase reliability and broaden usability by integrating 3D models of humans, mines, and marine life into diverse environmental settings for enhanced information richness. Additionally, by incorporating advanced techniques such as diffusion models and generative AI, we can introduce greater diversity and improve the representativeness of the generated datasets. Moving forward, addressing these limitations will be crucial for enhancing the reliability and scalability of simulation-based approaches in sonar image analysis. We also envisage refining our framework's flexibility through multi-object detection and tracking. We anticipate that the S3Simulator dataset will significantly advance AI technology for marine exploration and surveillance by offering valuable sonar imagery for research purposes.
\textcolor{black}{\section{Acknowledgments}
This work was partially supported by the Naval Research Board (NRB), DRDO, Government of India under grant number: NRB/505/SG/22-23.}



\vspace{-.2cm}
\bibliographystyle{splncs04}
\bibliography{Reference}
\end{document}